\documentclass[journal,twoside,web]{ieeecolor}

\usepackage{lcsys}
\usepackage{cite}
\usepackage{amsmath,amssymb,amsfonts}
\usepackage{algorithmic}
\usepackage{graphicx}
\usepackage{textcomp}
\usepackage{tabularx}
\usepackage[normalem]{ulem}
\usepackage{dsfont}
\usepackage{graphicx}
\usepackage{subcaption}
\usepackage{algorithm}
\usepackage{algorithmic}
\usepackage{booktabs}

\usepackage{dashrule}
\usepackage{orcidlink}

\let\labelindent\relax
\usepackage{enumitem}

\usepackage{tabularx}

\def\R{\mathbb{R}}
\def\T{[0,2\pi]}

\def\notes#1{\marginpar{\tiny #1}\typeout{Notes!
Notes!
Notes!
}}
\renewcommand{\notes}[1]{\typeout{notes!}}

\def\Expect{{\mathbb E}}

\def\R{\mathbb{R}}

\def\N{\mathcal{N}}

\def\beq{\begin{eqnarray}} 
\def\bc{\begin{center}} 
\def\be{\begin{enumerate}}
\def\bi{\begin{itemize}} 
\def\bs{\begin{small}}
\def\bS{\begin{slide}}
\def\ec{\end{center}} 
\def\ee{\end{enumerate}}
\def\ei{\end{itemize}}
\def\es{\end{small}}
\def\eS{\end{slide}}
\def\eeq{\end{eqnarray}}

\newcommand{\ud}{\,\mathrm{d}}

\newcounter{rmnum}

\newcounter{anum}

\renewcommand{\T}{T}

\def\BibTeX{{\rm B\kern-.05em{\sc i\kern-.025em b}\kern-.08em
    T\kern-.1667em\lower.7ex\hbox{E}\kern-.125emX}}
\begin{document}

\title{Generative models for simulation based filtering: \\
Formulations and Empirical Comparisons}

\author{Mohammad Al-Jarrah$^{\star,\dagger}$\orcidlink{0009-0006-0433-9230} \and
Wei Deng$^\ddagger\orcidlink{0000-0002-9655-8281}$ 
\and
Bamdad Hosseini$^\dagger$\orcidlink{0000-0001-5053-6223} \and 
Amirhossein Taghvaei$^\star$\orcidlink{0000-0002-1536-892X}
    {\thanks{$^\star$Department of Aeronautics \& Astronautics, University of Washington, Seattle; {\tt\small mohd9485@uw.edu,amirtag@uw.edu}.}} 
    {\thanks{$^\dagger$Department of Applied Mathematics, University of Washington, Seattle
        {\tt\small mohd9485@uw.edu,bamdadh@uw.edu}.}}
    {\thanks{$^\ddagger$ ML Research, Morgan Stanley
        {\tt\small weideng056@gmail.com}.}}
    \thanks{
    Part of this work was carried out while Mohammad Al-Jarrah was a ML Research intern at Morgan Stanley.
Mohammad Al-Jarrah and Amirhossein Taghvaei are supported by the National Science Foundation (NSF) award EPCN-2318977. Bamdad Hosseini is supported by the NSF award DMS-233767.}
}

\newif\iftodo \todotrue
\iftodo
  \newcommand{\todo}[1]{\textcolor{red}{[TODO: #1]}}
  \newcommand{\todonum}{\textcolor{red}{$\bullet$}}
\else
  \newcommand{\todo}[1]{}
  \newcommand{\todonum}{}
\fi

\newcommand{\dashedrule}{%
  \noindent\makebox[\linewidth]{%
    \leaders\hbox to 6pt{\hss\rule[0.6ex]{3pt}{0.5pt}\hss}\hfill}}

\maketitle
\thispagestyle{empty}

\begin{abstract}
This letter presents a unified formulation and a controlled numerical comparison of generative-model approaches to the nonlinear filtering problem. Under this formulation the analysis step is realized by a transport of the forecast distribution to the posterior, the approaches differing only in how that transport is selected and learned. We derive three new filters, based on stochastic interpolants, their deterministic flow-matching limit, and Schr\"odinger bridges realized through forward--backward SDEs. We develop a two-stage tuning procedure that separates the training of the generative model from its online refinement. The resulting methods are compared against the optimal transport filter (OTF), the Knothe--Rosenblatt filter (KRF), the sequential importance resampling (SIR) particle filter and the ensemble Kalman filter (EnKF), in terms of accuracy, computational time, and sensitivity to ensemble size and state dimension. The results indicate that every generative filter resolves multimodal posteriors that the EnKF and SIR do not, that no single generative framework dominates, the preferred method being set by the available online budget and ensemble size, and that the filters differ in the regularity of the particle trajectories they produce.
\end{abstract}


\section{INTRODUCTION}\label{sec:intro}

\IEEEPARstart{D}{eep} generative models have proven effective at sampling from complex,
high-dimensional distributions across a wide range of applications~\cite{bond2021deep}.
A structurally similar problem arises in
nonlinear filtering, where the object of interest is the posterior
distribution of the state of a dynamical system given a sequence of
noisy observations, and where an algorithm must represent that
posterior by a finite collection of samples. What distinguishes the
filtering setting is its recursive nature: the posterior evolves in
time, and samples must be produced online as each observation
arrives. This paper develops the application of several generative
modeling frameworks to this problem and compares them
empirically.

The use of generative models for filtering is motivated by a recent body of
work on coupling- or transport-based filters, in which the analysis
(conditioning) step is realized by a generative model that transports the forecast
distribution to the posterior~\cite{reich2013nonparametric,daum2012particle,elmoselhy2012bayesian,yang2016,spantini2022coupling,taghvaei2022optimal,al-jarrah2024nonlinear}.
Two constructions have received particular attention: the 
Knothe--Rosenblatt (KR) rearrangement filter (KRF)~\cite{spantini2022coupling}, and the optimal transport filter
(OTF)~\cite{al2023optimal,al-jarrah2024nonlinear}. Underlying both is the same
structural object: a block-triangular map $(y,x)\mapsto(y,T(y,x))$ that
transports the independent state-observation coupling 
to their joint law~\cite{kovachki2020conditional}. The two differ only in how such a map is 
represented and estimated.

In this paper we exploit the same triangular structure to
re-purpose several modern generative modeling frameworks for the
filtering task, including conditional flow matching~\cite{lipman2023flow}, the stochastic interpolant  
generalization~\cite{albergo2025}, and Schr\"odinger bridges 
~\cite{chen2022likelihood}. Beyond the development of these methods, our aim is a complete empirical study: we compare the
resulting filters in terms of accuracy, computational cost,
robustness, and the effort required to tune and implement them.
Generative approaches
to data assimilation have appeared recently
in~\cite{bao2024ensemble,transue2025flow}; these
works, 
however, 
develop a single mechanism in isolation, and a
controlled comparison 
under a common
transport formulation, identical ensembles, and matched computational
budgets is, to the best of our knowledge, not available. 
This gap
motivates 
the present 
study.

The outline and summary of contributions in the paper are as 
follows: 
Section~\ref{sec:problem} formulates the filtering problem and states the conditioning problem that the analysis step must solve at each observation time. Section~\ref{sec:methods} presents the triangular transport formulation under which a generative model
can be adapted towards conditioning tasks 
and derives three filters as instances of
that template: the stochastic interpolant filter (SIF), its
deterministic flow-matching specialization (FMF), and the
Schr\"odinger bridge filter (SBF), summarized in Table~\ref{tab:methods}.
Section~\ref{sec:numerics}
develops a two-stage tuning procedure specifically designed for
generative filtering models which separates the calibration of the
learned generative model from the subsequent calibration of the
filtering update and provides a common reproducible methodology for
selecting the algorithmic parameters of each filter.
This section also reports a controlled numerical comparison
against four established alternatives, namely OTF, KRF,
the ensemble
Kalman filter (EnKF)~\cite{evensen1994sequential}, and the sequential importance resampling (SIR)~\cite{gordon1993novel}
particle filter, on two benchmarks. 
Section~\ref{sec:discussion} compares the cost-accuracy trade-off of
the filters and their sensitivity to the problem dimension and
ensemble size, points to subtler differences between them, and closes
with directions for future work.

\section{PROBLEM FORMULATION}\label{sec:problem}

Consider the stochastic dynamical system
\begin{equation}\label{eq:hmm}
  X_t \sim a(\cdot \mid X_{t-1}), \quad X_0 \sim \pi_0, \qquad
  Y_t \sim h(\cdot \mid X_t),
\end{equation}
where $X_t \in \R^{n}$ is the hidden state, $Y_t \in \R^{m}$ is the
observation, $\pi_0$ is the initial distribution of the state,
$a(x' \mid x)$ is the transition kernel of the dynamics, and
$h(y \mid x)$ is the likelihood of the observation given the state.
The object of interest is the posterior distribution
$\pi_t := P_{X_t|Y_1,\ldots,Y_t}$, which
admits the two-step recursion
\begin{equation}\label{eq:recursion}
  \pi_{t|t-1} = \mathcal{A}[\pi_{t-1}], \qquad
  \pi_t = \mathcal{B}_{Y_t}[\pi_{t|t-1}].
\end{equation}
The propagation operator $\mathcal{A}$ transports the distribution
through the dynamics and the conditioning operator $\mathcal{B}_{y}$
applies Bayes' rule. For a distribution $\pi$ admitting a density on
$\R^{n}$, they are defined as
\begin{subequations}\label{eq:operators}
\begin{align}
  \text{(propagation)}\quad
  \pi \mapsto \mathcal{A}[\pi]
  &:= \int_{\R^{n}} a(\cdot \mid x)\, \pi(x)\,\mathrm{d}x,
  \label{eq:propagation} \\
  \text{(conditioning)}\quad
  \pi \mapsto \mathcal{B}_{y}[\pi]
  &:= \frac{h(y \mid \cdot)\, \pi(\cdot)}
           {\int_{\R^{n}} h(y \mid x)\, \pi(x)\,\mathrm{d}x}.
  \label{eq:conditioning}
\end{align}
\end{subequations}

A transport-based numerical realization of~\eqref{eq:recursion}
represents $\pi_t$ by an ensemble of $N$ particles
$\{X^1_t,\ldots,X^N_t\}$, propagated through the dynamics and then
updated by a map $\T_t : \R^{m}\times\R^{n} \to \R^{n}$:
\begin{subequations}\label{eq:particle}
\begin{align}
  X^i_{t|t-1} &\sim a(\cdot \mid X^i_{t-1}),
  \label{eq:particle-forecast} \\
  X^i_t &= \T_t\bigl(Y_t, \, X^i_{t|t-1}\bigr),
  \label{eq:particle-analysis}
\end{align}
\end{subequations}
for $i = 1,\ldots,N$. The map $\T_t$ is allowed to depend on auxiliary
randomness, in which case~\eqref{eq:particle-analysis} is a stochastic
update. The posterior is approximated by the empirical distribution
$\pi_t \approx \frac{1}{N}\sum_{i=1}^{N} \delta_{X^i_t}$, where
$\delta_x$ is the Dirac measure at $x$; in contrast to particle
filters, this ensemble is uniformly weighted. The forecast
step~\eqref{eq:particle-forecast} requires only the ability to
simulate the dynamics. The difficulty lies in the analysis
step~\eqref{eq:particle-analysis}, and specifically in identifying an
accurate map $\T_t$. We state this problem on its own below, dropping
the time index for clarity.

\noindent
\textbf{Problem statement.}
Let $\{X^i\}_{i=1}^{N}$ be independent samples from the forecast
distribution $\pi$, and let $Y^i \sim h(\cdot \mid X^i)$ be an
observation simulated for each ensemble member. The goal is to
construct a map $\T$ that reproduces the conditioning
operator~\eqref{eq:conditioning}:
\begin{align*}
  \text{Given:}\quad & \bigl(Y^i,\,X^i\bigr)_{i=1}^{N},
    \quad X^i \sim \pi, \quad Y^i \sim h(\cdot \mid X^i), \\
  \text{Find:}\quad & \textit{a map }
    \T : \R^{m}\times\R^{n} \to \R^{n} \textit{ such that }\\
  & \T(y,\cdot)_{\#}\,\pi = \mathcal{B}_{y}[\pi],
    \qquad \forall\, y \in \R^{m},
\end{align*}
where $\#$ denotes the push-forward operator. 
We refer to the pairs $\bigl(Y^i,X^i\bigr)_{i=1}^{N}$ as the
\emph{training data}. Their generation is
\emph{likelihood-free}, in the sense that it requires no evaluation of
the likelihood $h$, but only the ability to sample from it.

\section{GENERATIVE FILTERS}\label{sec:methods}

We begin by presenting the common triangular structure of  generative filters studied in this paper
and specify the characteristics of individual methods.

\subsection{Conditioning with triangular maps}\label{sec:transport}
The unifying element is that the analysis
step~\eqref{eq:particle-analysis} is realized by a \emph{triangular}
map, learned from the training data with a generative modeling
framework.
We say that a map $G:\R^m \times \R^n \to \R^m \times \R^n$ is triangular if there exists a map $T:\R^m \times \R^n \to \R^n$ such that 
\begin{align}\label{eq:triangular-map}
    G (y,z) = (y,T(y,z)),\quad \forall (y,z) \in \R^m \times \R^n.  
\end{align}
The signature of the triangular map is that it holds the $y$ components fixed\footnote{Although in a more general setting, not considered in this paper, the $y$ component is allowed to change according to any $y$-dependent map.} 
The triangular maps are useful for conditioning because of the following
elementary fact: for any joint distribution $P_{Y,X}$ on
$\R^{m}\times\R^{n}$ and any reference distribution $P_Z$ on $\R^{n}$, the following holds~\cite{kovachki2020conditional,al-jarrah2024nonlinear}:
\begin{align}\label{eq:triangular-map-conditioning}
 G_\# (P_Y \otimes P_Z) = P_{Y,X}\quad \Rightarrow \quad T(y,\cdot)_\# P_Z = P_{X|Y=y}. 
\end{align}

This suggests a recipe for re-purposing a generative 
model to solve the conditioning problem posed in
Section~\ref{sec:problem}: (i)  let the \emph{target} $P_{Y,X}$ be the joint distribution of
    the training data $\bigl(Y^i,\,X^i\bigr)_{i=1}^{N}$; (ii) let the \emph{source} be the distribution of the pairs
    $\bigl(Y^i, \, \overline X^i\bigr)_{i=1}^{N}$, where the
    $\overline{X}^i$ are obtained by randomly permuting the $X^i$, so that the source is the
    product measure $P_Y \otimes P_X$; 
    (iii) train a generative model to learn a (deterministic or stochastic) map of the triangular
    form~\eqref{eq:triangular-map} that transports the source to the
    target.
Any such map satisfies the hypothesis
of~\eqref{eq:triangular-map-conditioning}, and its second block
consequently satisfies
\begin{equation}\label{eq:consistency}
  T(y,\, \cdot)_{\#}\,P_X
  = P_{X\mid Y=y}
  = \mathcal{B}_y\bigl[\pi\bigr],
  \qquad \forall\, y \in \R^{m}.
\end{equation}

The condition~\eqref{eq:consistency} does not specify $T$ uniquely, and the
methods presented here differ precisely in how a
particular solution is selected and learned.

\subsection{Optimal Transport Filter (OTF)}\label{sec:otf}

OTF~\cite{al-jarrah2024nonlinear} selects $\T(y,\cdot)$ 
to be the optimal transport (OT) map from $P_X$ to  $P_{X|Y=y}$ for any observation value $y$. If the OT cost is quadratic, the OT map is obtained as the 
solution of the max-min problem~\cite[eq~(8c)]{al-jarrah2024nonlinear}
\begin{equation}\label{eq:maxmin}
\begin{aligned}
  \max_{f}\,&\min_{\T}\;\; J(f,T)
  := \Expect\bigl[ f(Y,X)\bigr] \\
  &+ \Expect\Bigl[ \tfrac{1}{2}\bigl\| \T(Y,\bar X) - \bar X \bigr\|^2
  - f\bigl(Y,\T(Y,\bar X)\bigr) \Bigr],
\end{aligned}
\end{equation}
where the first expectation is taken with respect to the joint law
$P_{Y,X}$ and the second with 
respect to the independent coupling
$P_Y \otimes P_X$.
In a numerical implementation, 
both $f$ and $\T$ are parameterized
by residual networks and trained adversarially.

\subsection{Flow Matching Filter
(FMF)}\label{sec:fmf}
In FMF the map $T$ is realized as the time-one flow of an ordinary
differential equation (ODE)~\cite{lipman2023flow}: for
each $(y,x) \in \R^{m}\times\R^{n}$, we set $T(y,x) = \xi_1$, where
\begin{equation}\label{eq:fmfode}
  \frac{\ud}{\ud \tau}\xi^i_\tau
  = b_\theta\bigl(y,\, \xi^i_\tau,\, \tau\bigr)\,\quad \xi^i_0=x,
\end{equation}
 and $b_\theta : \R^m \times \R^n \times [0,1] \to
\R^n$ is a learned vector-field. The artificial time $\tau \in [0,1]$ is distinct
from the filtering time $t$. The observation $y$ enters
\eqref{eq:fmfode} as a parameter and is held fixed along the flow, so
the associated map is triangular in the sense
of~\eqref{eq:triangular-map} by construction. The vector-field is obtained as the solution to the regression problem:
\begin{equation}\label{eq:fmfloss}
  \min_{\theta}\;
  \Expect\Bigl[
  \bigl\| b_\theta(Y,\, X_\tau,\, \tau) - \partial_\tau X_\tau \bigr\|^2
  \Bigr],
\end{equation}
where the expectation is taken over $\tau \sim \mathrm{Unif}[0,1]$,
$(Y,X) \sim P_{Y,X}$, and $\overline{X} \sim P_X$ drawn independently
of $Y$, and where
\begin{equation}\label{eq:fmfinterpolant}
  X_\tau = (1-\tau)\,\overline{X} + \tau X, \qquad \tau \in [0,1],
\end{equation}
is the linear interpolant between the source and the target. Its
derivative $\partial_\tau X_\tau = X - \overline{X}$ is available in
closed form. In practice, $\theta$ is obtained by minimizing the empirical
counterpart of~\eqref{eq:fmfloss} over the training data, and the analysis step is carried out by
integrating~\eqref{eq:fmfode} numerically from $\xi_0 = X^i$
with the observation fixed at $y = Y_t$.
\subsection{Stochastic Interpolant Filter 
(SIF)}\label{sec:sif}

The SIF~\cite{albergo2025} generalizes FMF along two independent axes: the
interpolant connecting source to target may carry noise, and the
generation may be carried out by a stochastic differential equation (SDE)
rather than an ODE.

The linear interpolant~\eqref{eq:fmfinterpolant} is generalized to
\begin{equation}\label{eq:sifinterp}
  X_\tau = (1-\tau)\,\overline{X} + \tau X + \gamma(\tau)\,Z,
  ~ \gamma(\tau) = \varepsilon\sqrt{\tau(1-\tau)},
\end{equation}
where $Z \sim \N(0,I_n)$ is independent of $(\overline{X}, X, Y)$ and
$\varepsilon \ge 0$ sets the noise level.  The case $\varepsilon=0$ recovers the linear interpolant~\eqref{eq:fmfinterpolant}. 
The velocity field is learned exactly as in~\eqref{eq:fmfloss}, with
the regression target now $  \partial_\tau X_\tau = X - \overline{X} + \dot\gamma(\tau)\,Z$. 
A second network $\eta_\theta$ is trained to obtain the score function,
\begin{equation}\label{eq:sifloss}
  \min_{\theta}\;
  \Expect\Bigl[
    \bigl\| \eta_\theta(Y, \, X_\tau,\, \tau) - Z \bigr\|^2
  \Bigr].
\end{equation}

Given the score, the deterministic flow~\eqref{eq:fmfode} may be
replaced by the SDE
\begin{equation}\label{eq:sifsde}
  \mathrm{d}\xi_\tau
  = \overline{b}_\theta\bigl(y,\, \xi_\tau,\, \tau\bigr)\,\mathrm{d}\tau
  + \lambda \gamma(\tau)\,\mathrm{d}W_\tau,
  \qquad \xi_0 = x,
\end{equation}
with drift
$ \overline{b}_\theta(y,x,\tau)
  = b_\theta(y,x,\tau) + \frac{\lambda^2 \gamma^2 (\tau)}{2}\,\eta_\theta(y,x,\tau)
$ where $\lambda\geq 0$ is a parameter that controls the diffusion coefficient.
The
score correction in~\eqref{eq:sifsde} compensates exactly for the
injected noise, so that~\eqref{eq:sifsde} and~\eqref{eq:fmfode}
share the same time-$\tau$ marginals for every choice of $\lambda$;
setting $\lambda \equiv 0$ recovers the ODE.

These generalizations, along with FMF itself, yield three filters that we implement
and compare in Section~\ref{sec:numerics}:
\begin{enumerate}[label=(\arabic*), leftmargin=2.2em]
  \item[(1)] FMF: \eqref{eq:fmfode} with deterministic interpolant
    ($\varepsilon = 0$, $\lambda=0$);
  \item[(2)] SIF-ODE: \eqref{eq:fmfode} with stochastic interpolant
    ($\varepsilon > 0$, $\lambda=0$);
  \item[(3)] SIF-SDE: \eqref{eq:sifsde} with  stochastic interpolant   ($\varepsilon > 0$, $\lambda=1$);
\end{enumerate}

\subsection{Schr\"odinger Bridge Filter (SBF)}\label{sec:sbf}
As in SIF, in SBF the map $T$ is realized as the time-one flow
of a stochastic differential equation,
\begin{align*}
    &\ud \xi_\tau = u_\theta(Y,\, \xi_\tau,\, \tau) \ud \tau + g \ud W_\tau\\
    \text{s.t.}\quad &(Y,\, \xi_0) \sim P_Y \otimes P_X, ~ (Y,\xi_1) \sim P_{Y,X}
\end{align*}
where $g > 0$ is the diffusion coefficient. 
What distinguishes
SBF is how the drift is selected: among all drifts that meet the two
marginal constraints, $u_\theta$ is the one of minimum energy,
\begin{equation}\label{eq:sbfcontrol}
  \min_{u}\;
  \Expect\Bigl[\textstyle\int_0^1
    \tfrac{1}{2}\|u(Y,\, \xi_\tau,\, \tau)\|^2\,\ud\tau\Bigr].
\end{equation}

The contrast with FMF and SIF is worth
stating: In SIF and FMF, the coupling between source and target is prescribed
in advance by the interpolant, and learning
reduces to a regression against a closed-form target. 
In SBF the
coupling is an unknown to be determined by the optimality
condition~\eqref{eq:sbfcontrol}, with $g$ playing the role of an entropic regularization parameter (the limit $g=0$ yields the OT map). The resulting problem has no
closed-form solution and must be solved iteratively.

We compute $u_\theta$ using the forward--backward SDE framework
of~\cite[Alg.~3]{chen2022likelihood}. The method parameterizes the
forward and backward drifts by two neural networks and trains them
alternately against likelihood objectives derived from the
forward--backward SDE system with each update 
requiring simulation of trajectories under the current drift. This simulation-in-the-loop
requirement is the principal practical difference from FMF and SIF,
whose training is simulation-free, and it is one of the properties we
study in Section~\ref{sec:numerics}.

\subsection{Knothe--Rosenblatt
rearrangement Filter (KRF)}\label{sec:KRF}
In KRF~\cite{spantini2022coupling}, the map $T$ is obtained in two
steps: a triangular map to a Gaussian reference is learned by maximum
likelihood, and the analysis is then realized by composing that map
with its inverse evaluated at the realized observation.

For the first step, let $\mathcal{T}$ denote the class of monotone triangular maps
$S : \R^{m}\times\R^{n} \to \R^{n}$ of the form
\begin{equation*}
  S(y,x) = \bigl(
    S^{1}(y,x_1),\;
    S^{2}(y,x_1,x_2),\;\ldots,\;
    S^{n}(y,x_1,\ldots,x_n)
  \bigr),
\end{equation*}
whose $k$-th component is strictly increasing in $x_k$. Note that the
triangular structure here is with respect to the state coordinates:
each component depends on the full observation but only the first $k$
components of the state. A map $S \in \mathcal{T}$ is trained so that
$S(Y,X) \sim \N(0,I_n)$ independently of $Y$, by minimizing the
negative log-likelihood,
\begin{equation}\label{eq:KRF}
  \min_{S \in \mathcal{T}}\;
  \Expect\Bigl[
    \tfrac{1}{2}\bigl\|S(Y,X)\bigr\|^2
    - \textstyle\sum_{k=1}^{n} \log \partial_{x_k} S^{k}(Y,X)
  \Bigr],
\end{equation}
with the expectation approximated with the training data. Monotonicity
guarantees that the Jacobian determinant appearing
in~\eqref{eq:KRF} is positive.
In practice, we integrate a softplus-positive slope field with a small positive floor, which makes each component monotone in its own coordinate.

The analysis step then proceeds by mapping each particle to the
reference using its own simulated observation and mapping back using
the realized one:
\begin{equation}\label{eq:KRFcomposed}
  T(y,X^i)
  = S\bigl(y,\cdot\bigr)^{-1}
    \circ S\bigl(Y^i,\, X^i\bigr).
\end{equation}
The inverse
in~\eqref{eq:KRFcomposed} is not available in closed form 
but can be evaluated one coordinate at a time by a
bisection scheme and backward substitution.

\begin{table}[t]
\caption{Summary of the generative filters, Section~\ref{sec:methods}.}
\label{tab:methods}
\centering
\footnotesize
\setlength{\tabcolsep}{3pt}
\begin{tabularx}{\columnwidth}{@{}l
  >{\raggedright\arraybackslash}X
  >{\raggedright\arraybackslash}X
  >{\raggedright\arraybackslash}X@{}}
\toprule
Filter & Transport/coupling & Learning & Inference \\
\midrule
OTF~\cite{al-jarrah2024nonlinear}  & OT map $\T$         & adversarial~\eqref{eq:maxmin}             & one-shot \\
FMF  & det. interpolant& regression~\eqref{eq:fmfloss} & ODE, Euler steps \\
SIF  & stoch. interpolant & regression~\eqref{eq:sifloss}         & ODE, or SDE$^\dagger$ \\
SBF  & fwd/bwd drifts     & min. energy~\eqref{eq:sbfcontrol}                     &  (reversed) SDE$^\dagger$ \\
KRF~\cite{spantini2022coupling} & KR map $S$ & max.\ lik.~\eqref{eq:KRF}            & one-shot, bisection \\
\bottomrule
\end{tabularx}
\par\vspace{0.4ex}
\parbox{\columnwidth}{\scriptsize $^\dagger$Stochastic inference:
repeated evaluation at a fixed observation yields different
particles.}
\end{table}
\section{NUMERICAL RESULTS}\label{sec:numerics}

This section compares the generative filters of
Section~\ref{sec:methods} on 
a quadratic observation
model and the Lorenz-63 system. The code used to produce the results
is available
online.\footnote{\url{https://github.com/Mohd9485/Generative_Filters}}

\subsection{Tuning procedure}\label{sec:tuning}
We develop a tuning procedure to select the hyperparameters of each
generative filtering algorithm. These hyperparameters include (i) the
network width and depth to parameterize transport maps or vector fields; (ii) learning rate, batch size, iteration number for training; (iii) where applicable, parameters controlling the
inference procedure, such as the step size used to integrate
the ODE or SDE. We use Bayesian optimization~\cite{lindauer2022smac3}
to search over these hyperparameters. The tuning procedure is divided
into two stages to account for the two training regimes encountered
during filtering. 

\textbf{Stage 1: cold spin-up and weight initialization.} The first stage tunes the hyperparameters used to
train the networks from scratch at the initial analysis step $t=1$.  Because a single analysis step is
simulated, each evaluation is computationally
inexpensive, allowing the hyperparameter search to be carried out over
a sufficiently large number of training iterations. The best-performing
configuration is then retrained, and the resulting network weights are
stored and used as the common initialization for the second stage.

\textbf{Stage 2: online refinement.} The
second stage tunes only the optimizer hyperparameters used to refine the networks at
subsequent analysis steps. Here we warm-start the networks at 
the weights from previous steps which allows us to perform 
smaller training runs with fewer hyperparameters.
The filtering performance is evaluated over
several consecutive analysis steps, so that the effect of these
hyperparameters on the recursive filtering procedure is captured.
The tuning is repeated at several
\emph{budget levels}, where each level specifies the number of optimizer
iterations permitted per analysis step, from zero up to a method-dependent maximum chosen to keep the total computational
tuning cost manageable.

\subsection{Error metric and sampling floor baseline}
Accuracy is quantified by the axis-aligned sliced Wasserstein-$2$ (aa-SW2)~\cite{al-jarrah2025error},
\begin{equation}\label{eq:aasw2}
  \mathcal{S}(\hat\pi_t, \tilde\pi_t)
  := \frac{1}{n}\sum_{k=1}^{n}
  W_2\bigl( (e_k)_\#\hat\pi_t,\; (e_k)_\#\tilde\pi_t \bigr),
\end{equation}
where $(e_k)_\#$ is the
push-forward onto the $k$-th coordinate, $\hat\pi_t$ the filter
ensemble, and $\tilde\pi_t$ a reference posterior. Each summand is a
one-dimensional OT problem, evaluated from the two
empirical quantile functions.

For the three-dimensional Lorenz-63 model in Section~\ref{sec:l63}, the reference $\tilde\pi_t$ is obtained from SIR with a large number of particles $N=10^6$. 
For the quadratic observation model in Section~\ref{sec:dynamic}, SIR suffers from the curse of dimensionality and does not provide a reliable reference. 
To resolve this, we exploit the independent block structure
of the model~\eqref{eq:dynamic} and run independent SIR filters per block 
(two-dimensional) with
$N = 10^5$ particles each, and stack the resulting
marginals.

The aa-SW2 distance between finite empirical measures contains an irreducible sampling error component. We estimate this \emph{sampling floor} by drawing an independent sample of size $N$ from the same reference posterior and computing its aa-SW2 distance to $\tilde{\pi}_t$ and time averaging. This quantifies the error induced solely by the finite ensemble size. Consequently, no  algorithm is expected to achieve errors below this level.

\subsection{The quadratic
observation model}\label{sec:dynamic}

Consider the system
\begin{equation}\label{eq:dynamic}
  X_t = F X_{t-1} + \sigma V_t, \qquad
  Y_t = H(X_t) + \gamma W_t ,
\end{equation}
where $\{V_t\}$ and $\{W_t\}$ are mutually independent sequences of
i.i.d.\ standard Gaussian random vectors. The matrix $F$ is
block diagonal, assembled from $m$ decoupled two-dimensional blocks, each representing a stable oscillatory dynamic. 
The observation map is quadratic 
$H(x) = (x_1^2,\, x_3^2,\, \ldots,\, x_{2m-1}^2)$ 
and  acts on the first
coordinate of each block, so that exactly one coordinate per block is
observed directly. 
The remaining parameters are $X_0 \sim \N(0, I_n)$,
$\sigma^2 = \gamma^2 = 10^{-1}$, and 
$T = 20$. 
Although the dynamics are linear, the quadratic observation leaves the sign of each observed
coordinate unresolved, resulting in a bimodal posterior along
the entire trajectory.

\begin{figure*}[t]
\centering

\begin{subfigure}[b]{\linewidth}
  \centering
  \includegraphics[width=\linewidth]%
    {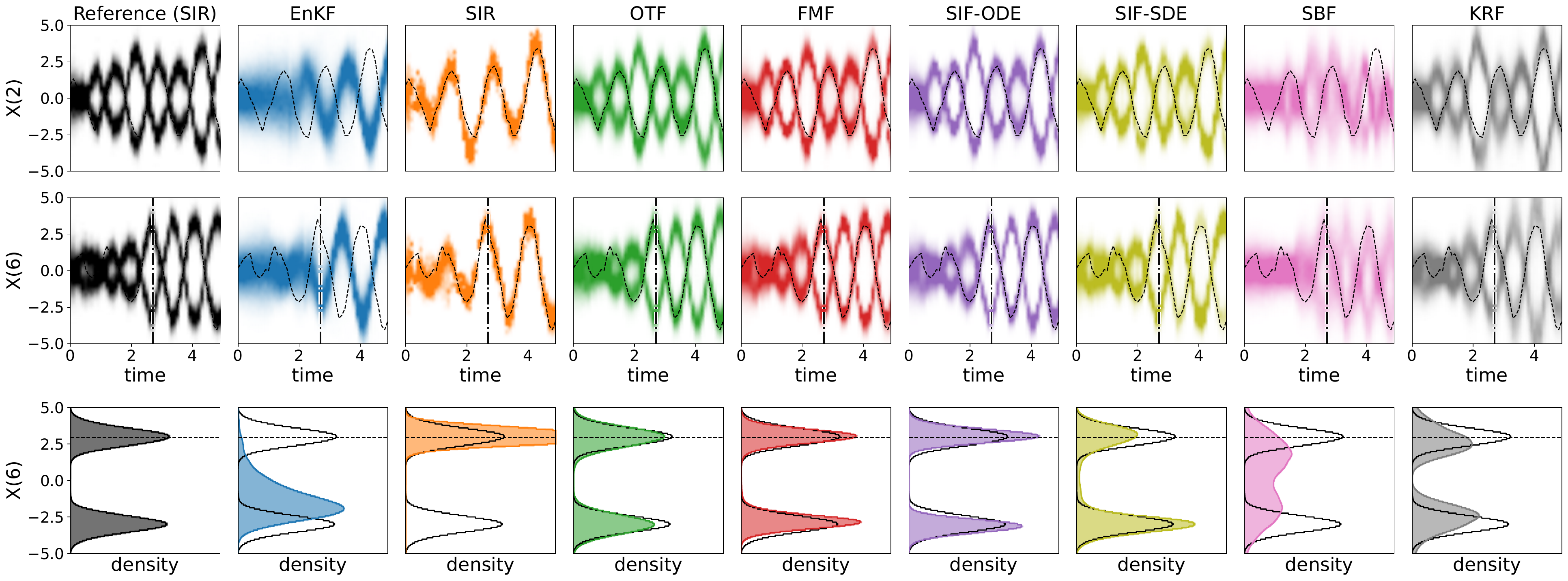}
  \caption{Quadratic benchmark~\ref{sec:dynamic}, $n = 10$,
  $N = 10^4$. Upper two rows: two of the unobserved coordinates over
  the horizon. Lower row: the marginal density of the second of these
  at $t = 2.7$, marked by the vertical line above, with the reference
  drawn in outline on every panel.}
  \label{fig:density-quad}
\end{subfigure}

\vspace{0.5ex}
\hdashrule[0.5ex]{\linewidth}{0.6pt}{3pt 2pt}
\vspace{0.0ex}

\begin{subfigure}[b]{\linewidth}
  \centering
  \includegraphics[width=\linewidth]%
    {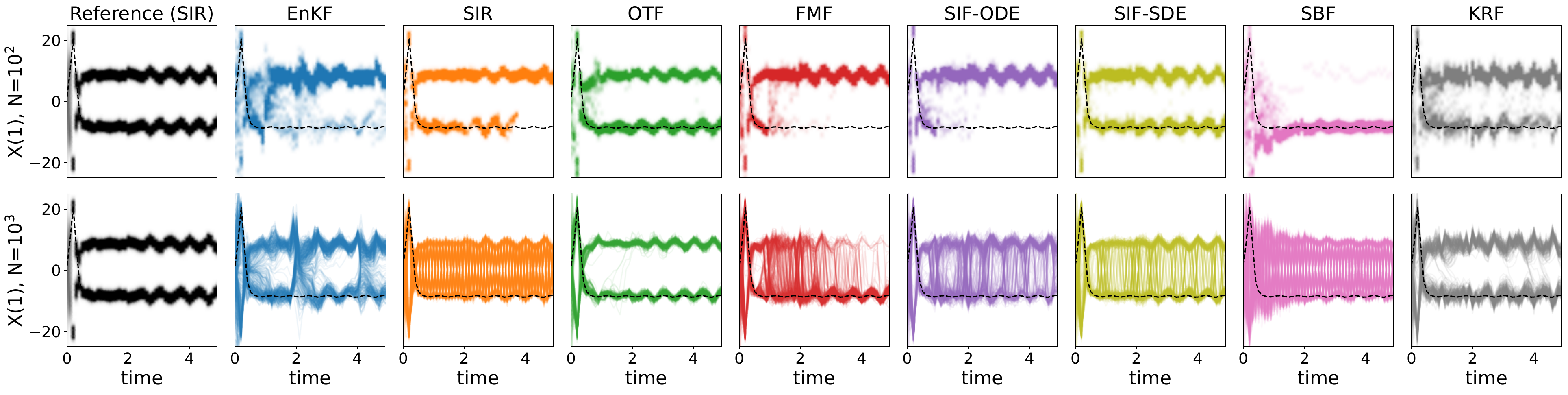}
  \caption{Lorenz-63 benchmark~\ref{sec:l63}, first coordinate, at ensemble sizes
  $N = 10^2$ (upper row) and $N = 10^3$ (lower row).}
  \label{fig:density-l63}
\end{subfigure}

\caption{Marginal posterior densities produced by each filter on the
two benchmarks. In both panels the leading column is the reference
posterior obtained from the SIR filter with a large particle count,
and the dashed line is the true state. Within each panel all methods
share the color scale and the vertical range, and the densities are
not rescaled.}
\label{fig:density}
\end{figure*}

Fig.~\ref{fig:density-quad} compares the marginal posterior densities
obtained by each method with the reference posterior. The first two rows
show the time evolution of the particle densities for two
unobserved states, together with the corresponding reference densities,
while the third row shows the posterior density of the second row at the time instant indicated by the dashed line.
It is observed that all generative filtering methods recover the bimodal structure of the reference posterior, with the exception of SBF. In contrast, the SIR particle filter
exhibits mode collapse, with the particle ensemble concentrating on
only one of the two posterior modes, while the EnKF fails to represent
the bimodal structure due to its underlying Gaussian
approximation. These qualitative results indicate that the generative filters
 provide a  more faithful representation of the
non-Gaussian posterior structure in this benchmark. 

\subsection{Scaling studies}
\begin{figure*}[ht]
\centering
\begin{subfigure}[b]{0.245\textwidth}
  \centering
  \includegraphics[width=\linewidth,trim={0 0 0 0},clip]{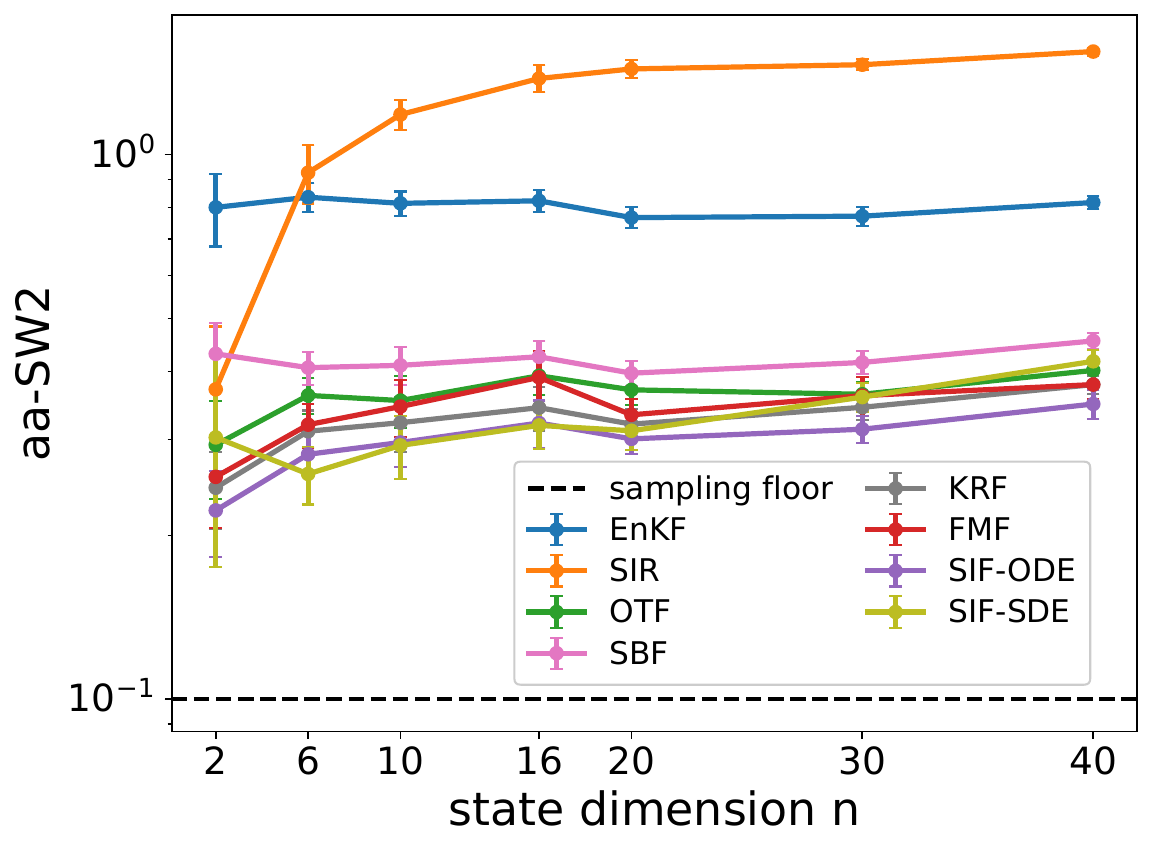}
  \caption{Error vs.\ dimension.}
  \label{fig:sw2_dim}
\end{subfigure}
\hfill
\begin{subfigure}[b]{0.245\textwidth}
  \centering
  \includegraphics[width=\linewidth,trim={0 0 0 0},clip]{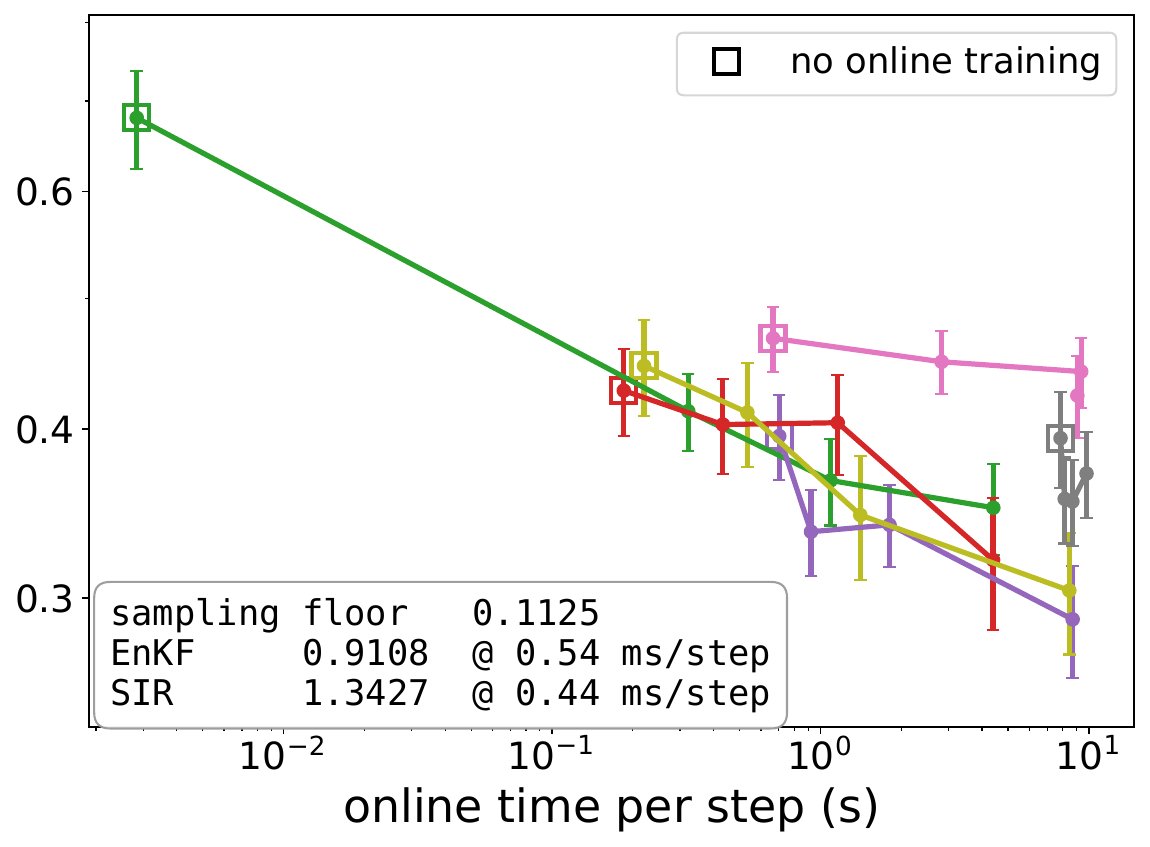}
  \caption{Error vs.\ online cost.}
  \label{fig:sw2_time}
\end{subfigure}
\hfill
\begin{subfigure}[b]{0.245\textwidth}
  \centering
  \includegraphics[width=\linewidth,trim={0 0 0 0},clip]{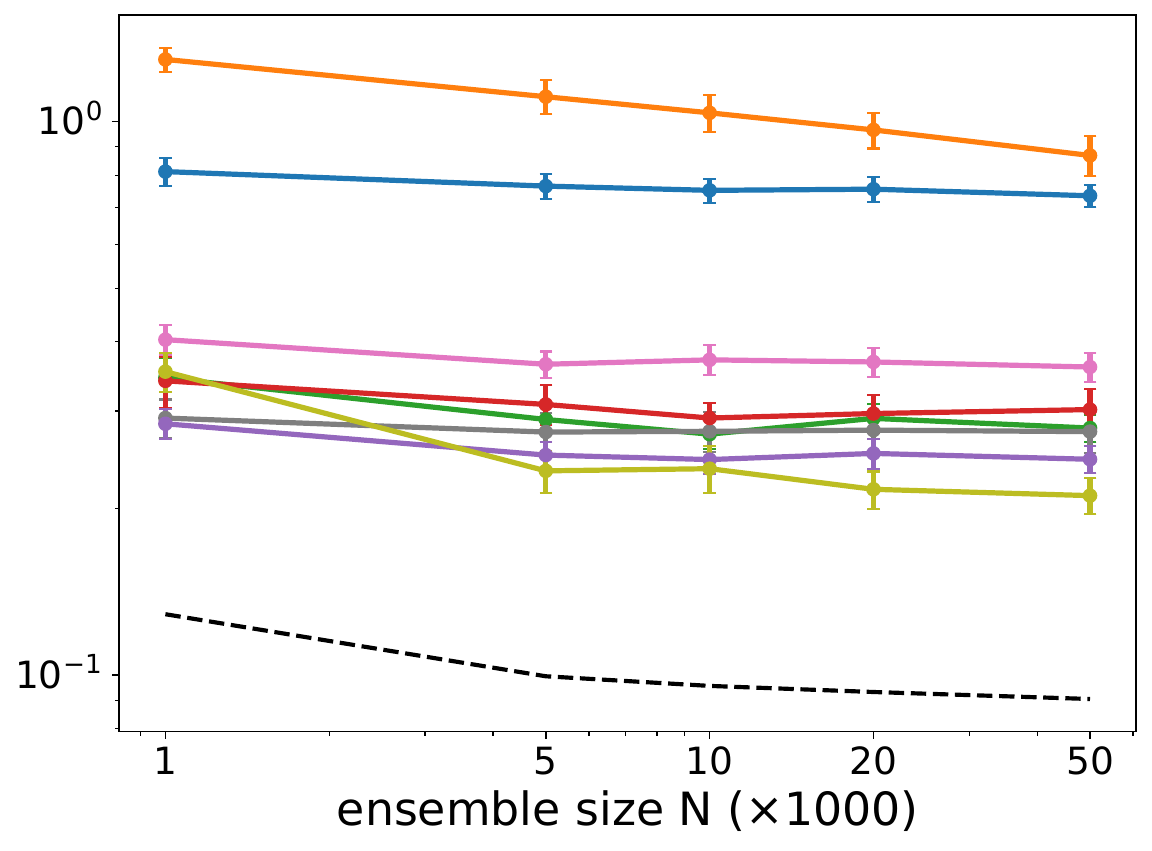}
  \caption{Quadratic~\ref{sec:dynamic}: Error vs.\ N.}
  \label{fig:sw2_N}
\end{subfigure}
\hfill
\begin{subfigure}[b]{0.245\textwidth}
  \centering
  \includegraphics[width=\linewidth,trim={0 0 0 0},clip]{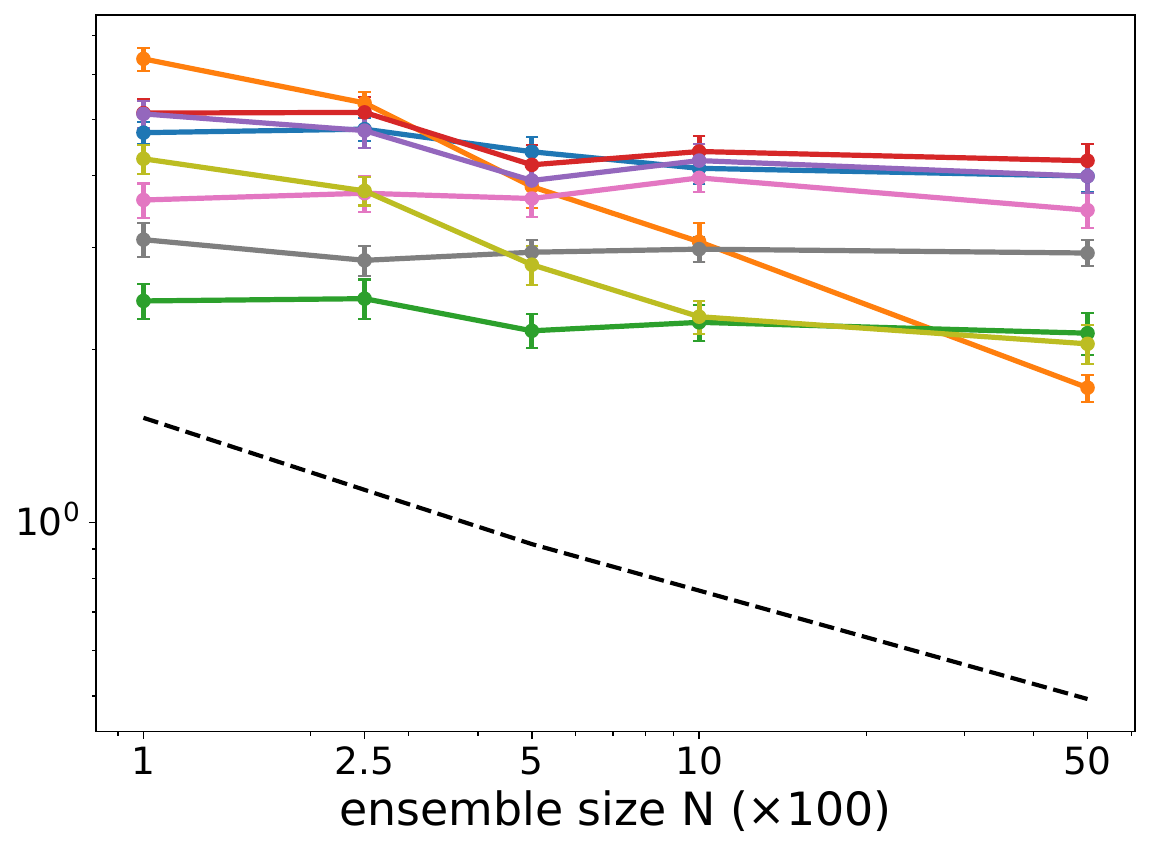}
  \caption{Lorenz-63~\ref{sec:l63}: Error vs.\ N.}
  \label{fig:sw2_N_L63}
\end{subfigure}
\caption{
Quantitative comparison on the benchmarks in
Sections~\ref{sec:dynamic} and~\ref{sec:l63}, averaged over
$T$ analysis steps and
$20$
independent simulations. Error bars denote the standard error of the
mean, and the dashed line denotes the sampling floor.
(a) Error versus state dimension.
(b) Error versus online refinement cost; the ringed point denotes zero
online refinement.
(c,d) Error versus ensemble size for the quadratic observation model
and Lorenz-63, respectively.
}
\label{fig:sweeps}
\end{figure*}

Three sweeps are reported in Fig.~\ref{fig:sweeps}~(a-b-c), designed to
isolate the dependence of the error upon the dimension of the state,
the resolution of the ensemble, and the computational budget expended
at each analysis step. The first varies the dimension
at fixed $N = 10^4$; the second varies
the ensemble size
at fixed $n = 10$; and the third varies the online budget
across the four levels at which the second tuning stage was
performed, at fixed $n = 10$ and $N = 10^4$.

In the dimension sweep Fig.~\ref{fig:sw2_dim} it is observed that the SIR filter degrades quickly, consistent with 
the known weight degeneracy, whereas the error of the
EnKF is nearly independent of $n$ and remains high throughout, as a
Gaussian ansatz applied to a bimodal posterior must.  
All generative filters lie well below both EnKF and SIR at every dimension, degrading only mildly
with $n$, and with SBF being the weakest method. 
In the budget sweep Fig.~\ref{fig:sw2_time} the OTF benefits most from online
refinement, its error falling by approximately a factor of two between
the zero budget and the maximum online time, 
whereas the SBF and the KRF benefit very little from additional online training; the SIF
configurations attain the lowest error at 
the highest budget, at a
cost some four orders of magnitude above that of 
the EnKF and the SIR filter.
In the ensemble-size
sweeps Fig.~\ref{fig:sw2_N} 
the error of every method depends only weakly upon $N$ and
remains well above the sampling floor across the range considered,
indicating that the residual error is dominated by the approximation
quality of the learned map rather than the ensemble size.
 
\subsection{The Lorenz-63 benchmark}\label{sec:l63}
We consider the three-dimensional Lorenz-63 system where only the third component of the state is observed. The observation model results in a bimodal posterior in the first two components due to the sign symmetry in the model. The model setup and its parameter values follow those
of~\cite{al-jarrah2025error}.

To test the robustness of each filter to its tuning, we apply here the
configurations selected for the quadratic benchmark of
Section~\ref{sec:dynamic}, without re-tuning for the present model.
Fig.~\ref{fig:density-l63} compares the marginal posterior densities
(first row) and corresponding particle trajectories (second row) for
the first coordinate at $N=10^2$ and $N=10^3$. At the smaller ensemble
size, several of the generative filters struggle to recover the
bimodal posterior, whereas the transport-based methods, OTF
and KRF, more consistently resolve both modes. The particle
trajectories in the second row further show that, for the non-transport-based
methods, particles frequently switch between the two modes across
successive analysis steps, whereas OTF and KRF tend to preserve the
mode occupied at the previous time step. This temporal persistence
may be advantageous for time-series inference, although its
implications for filtering performance 
remain subject to future investigation. Fig.~\ref{fig:sw2_N_L63} shows OTF and KRF achieve lower
aa-SW2 errors at small ensemble sizes, with OTF providing the
strongest improvement over the remaining methods.

\section{Discussion}\label{sec:discussion}
Among the new generative filters, FMF and the SIF variants exhibit
nearly the same cost--accuracy trade-off, with a slight advantage to
SIF-SDE that widens as the ensemble size grows. This is expected: these methods solve similar regression problems, and the noise carried by
the stochastic interpolant makes SIF-SDE more sample-hungry, so its
advantage is realized only once the ensemble is large. 
By contrast, SBF is more expensive because it does not
admit a simple regression loss and 
requires SDE simulations
within the training loop.

The two transport filters, OTF and KRF, are limited by different
mechanisms. OTF is constrained by the 
adversarial training of
\eqref{eq:maxmin}, whose convergence dominates the error.
KRF
is instead constrained at inference, where the map must be inverted by
bisection and back substitution per coordinate. 

Comparing across the two families, OTF and SIF-SDE trace similar
cost--accuracy curves: OTF is preferable when the online budget/
ensemble is small, and SIF-SDE otherwise.

A further qualitative difference is visible in the particle
trajectories: under OTF and KRF trajectories remain regular, reflecting the
smoothness of the underlying maps, whereas under FMF, SIF, and SBF the
paths of nearby particles criss-cross. What
this implies for downstream control and inference tasks which may be
sensitive to the regularity of particle trajectories
is left to future work.

\bibliographystyle{IEEEtran}
\bibliography{references}

\end{document}